\documentclass{article} % For LaTeX2e
\usepackage{iclr2026_conference,times}
\makeatletter
\def\iclrruler#1{}
\makeatother

\usepackage{amsmath,amsfonts,bm}

\def\eqref#1{equation~\ref{#1}}
\def\1{\bm{1}}

\DeclareMathAlphabet{\mathsfit}{\encodingdefault}{\sfdefault}{m}{sl}
\SetMathAlphabet{\mathsfit}{bold}{\encodingdefault}{\sfdefault}{bx}{n}

\usepackage{hyperref}
\usepackage{url}
\usepackage{graphicx}
\usepackage{booktabs}       % professional-quality tables
\usepackage[table]{xcolor}
\usepackage{enumitem}
\usepackage{algorithm}      % For the floating environment
\usepackage{algpseudocode}  % For the pseudocode style
\usepackage{amssymb}
\usepackage{multirow}

\newcommand{\ERPO}{\texttt{ERPO}}

\title{Explore Data Left Behind in Reinforcement Learning for Reasoning Language Models}

\author{
Chenxi Liu\textsuperscript{1,2}\thanks{Work done during an internship at ByteDance.},
Junjie Liang\textsuperscript{2},
Yuqi Jia\textsuperscript{2,3},
Bochuan Cao\textsuperscript{4},
Yang Bai\textsuperscript{2},
Heng Huang\textsuperscript{1},
Xun Chen\textsuperscript{2} \\
\textsuperscript{1}University of Maryland, College Park \quad
\textsuperscript{2}ByteDance \quad 
\textsuperscript{3}Duke University \quad \\
\textsuperscript{4}Pennsylvania State University \\
}

\arxivcopy
\begin{document}

\maketitle

\begin{abstract}
Reinforcement Learning with Verifiable Rewards (RLVR) has emerged as an effective approach for improving the reasoning abilities of large language models (LLMs). The Group Relative Policy Optimization (GRPO) family has demonstrated strong performance in training LLMs with RLVR. However, as models train longer and scale larger, more training prompts become residual prompts—those with zero-variance rewards that provide no training signal. Consequently, fewer prompts contribute to training, reducing diversity and hindering effectiveness. To fully exploit these residual prompts, we propose the \textbf{E}xplore \textbf{R}esidual Prompts in \textbf{P}olicy \textbf{O}ptimization (ERPO) framework, which encourages exploration on residual prompts and reactivates their training signals. ERPO maintains a history tracker for each prompt and adaptively increases the sampling temperature for residual prompts that previously produced all-correct responses. This encourages the model to generate more diverse reasoning traces, introducing incorrect responses that revive training signals. Empirical results on the Qwen2.5 series demonstrate that ERPO consistently surpasses strong baselines across multiple mathematical reasoning benchmarks.Source code is available at \url{https://github.com/DawnLIU35/ERPO}
\end{abstract}

% However, GRPO requires both correct and incorrect responses for each prompt to provide valid training signals. This requirement becomes increasingly difficult to satisfy as models grow larger and train longer, since they tend to generate more all-correct responses.

\section{Introduction}
\label{introduction}

Large language models (LLMs) have become the foundation of modern artificial intelligence, exhibiting strong performance across domains such as mathematics, programming, and scientific problem solving~\citep{team2023gemini,guo2025deepseek,yang2025qwen3,zheng2025parallel,chen2024your,chen2023gpt,chen2024watermark,shirkavand2025cost,shirkavand2025bilevel,chen2025enhancing}. A central factor behind these advancements is their capacity for extended reasoning, where models construct coherent, multi-step chains of thought to address complex tasks~\citep{wei2022chain,yao2023tree,muennighoff2025s1}. Reinforcement learning (RL) has emerged as a key paradigm for strengthening this capability, enabling LLMs to refine their responses through interaction-driven feedback and alignment with verifiable signals or human preferences~\citep{schulman2017proximal,ouyang2022training,rafailov2023direct,chen2024optune}. In particular, reinforcement learning with verifiable rewards (RLVR) has proven especially effective, as it leverages tasks with automatically checkable outcomes to provide reliable supervision for scaling reasoning abilities~\citep{shao2024deepseekmath,guo2025deepseek,yang2025qwen3,liu2025modality}.

Among recent advances in reinforcement learning for LLMs, Group Relative Policy Optimization (GRPO) has emerged as a widely adopted RLVR framework~\citep{shao2024deepseekmath,guo2025deepseek}. Building on this foundation, subsequent research has sought to address key issues of GRPO, including entropy collapse, reward noise, and training instability~\citep{yu2025dapo,cui2025entropy,zheng2025group}. Furthermore, as an on-policy algorithm, GRPO has motivated efforts to develop more effective sampling strategies beyond basic random decoding~\citep{xu2025not,zheng2025first,hou2025treerl}. 

In this work, we identify a limitation shared by the GRPO family of algorithms: as training steps and model size increase, more training prompts become residual prompts that no longer provide training signals yet still contain valuable information that can benefit model performance. Residual prompts are those that initially provide effective training signals at the beginning of training but eventually provide zero training signal or are filtered out by the RL algorithms as the well-trained policy generates all-correct responses for them. This reduces training diversity over time and ultimately hinders further improvement through RL. Furthermore, residual prompts retain learning potential that can be leveraged to further improve model performance, as they help the model retain acquired abilities and may yield novel reasoning traces. Moreover, residual prompts are not necessarily robust—small perturbations, such as increasing the sampling temperature, can easily induce errors. Table~\ref{tab:gen_temp} reports the proportion of residual prompts with all-correct responses in the training data under different sampling temperatures and model scales.

\begin{table}[t!]
\centering
\caption{Proportion of prompts with all-correct responses under different sampling temperatures and model scales. The proportion increases with RL training process and larger model sizes, leaving more residual prompts, thereby reducing diversity and wasting valuable training signals.}
\setlength{\tabcolsep}{4pt}
\renewcommand{\arraystretch}{1.2}
\begin{tabular}{lccc}
    \toprule
      & {\textbf{$T=1.0$}} & {\textbf{$T=1.1$}} & {\textbf{$T=1.2$}} \\
    \midrule[0.6pt]
    { Qwen2.5-3B }   & $0\%$ & -- & --  \\
    { Qwen2.5-3B + DAPO}   & $8.7\%$  & $6.2\%$ & $2.8\%$    \\
    { Qwen2.5-7B }   & $0\%$ & -- & --  \\
    { Qwen2.5-7B + DAPO}   & $21.3\%$ & $15.5\%$ & $5.5\%$    \\
    { Qwen2.5-32B }  & $0\%$ & -- & --  \\
    { Qwen2.5-32B + DAPO}  & $74.8\%$ & $62.1\%$ & $34.8\%$    \\
    \bottomrule
\end{tabular}
\vspace{-6pt}
\label{tab:gen_temp}
\end{table}

To better exploit the residual prompts left behind during training, we propose the \textbf{E}xplore \textbf{R}esidual Prompts in \textbf{P}olicy \textbf{O}ptimization (\ERPO{}) framework. \ERPO{} introduces a novel sampling strategy that maintains a history tracker for each prompt and adaptively increases the sampling temperature for residual prompts that have previously produced all-correct responses. Specifically, \ERPO{} records how many times a model generates all-correct responses for each prompt, and the sampling temperature is determined by this count. The more frequently a prompt yields all-correct responses, the higher the sampling temperature assigned to it, thereby encouraging greater exploration. As shown in Table~\ref{tab:gen_temp}, increasing the sampling temperature enables the model to explore more diverse reasoning traces and generate incorrect responses, which reactivates the training signal and alleviates the collapse of prompt diversity.

Overall, \textbf{our contributions} can be summarized as follows:

\begin{itemize}[leftmargin=2em]
    \setlength{\itemsep}{0.3em}
    \item We identify a key limitation of the GRPO family: residual prompts accumulate as training progresses and models scale, leading to reduced training diversity and the loss of valuable training signals from residual prompts.
    \item We propose the \ERPO{} framework, which encourages models to adaptively explore residual data and recover their learning potential. \ERPO{} maintains a history tracker for each prompt and adaptively increases the sampling temperature for residual prompts.
    \item Extensive experiments on several math reasoning benchmarks demonstrate the effectiveness of \ERPO{} in both average and majority-vote evaluations, with particularly strong improvements on data that are likely not contaminated, such as AIME2025.
\end{itemize}
\section{Related Work}
\label{related_work}

\textbf{Reinforcement learning for LLM reasoning.} Reinforcement learning (RL) has become a central approach for enhancing the reasoning abilities of large language models (LLMs) in domains such as mathematics, programming, and problem solving~\citep{dubey2024llama,zhou2025reinforcing}. Early general-purpose algorithms like Proximal Policy Optimization (PPO) provided a practical framework for fine-tuning LLMs through sampled rollouts and reward feedback~\citep{schulman2015trust,schulman2017proximal}. More recently, RLVR methods such as Group Relative Policy Optimization (GRPO) have emerged as effective alternatives to PPO, removing the critic model while maintaining strong performance on reasoning benchmarks~\citep{guo2025deepseek,shao2024deepseekmath}. Several extensions have been proposed to address the limitations of GRPO: \citet{cui2025entropy,wang2025beyond,cheng2025reasoning,zheng2025first} mitigates the entropy collapse problem during training; \citet{zheng2025group,yang2025dcpo} aims to stabilize the optimization process, and DAPO~\citep{yu2025dapo} tackles both issues while filtering noisy rewards for training data. However, all these methods obtain no training signal from residual prompts, thereby missing valuable information during training. To address this limitation, \ERPO{} reactivates the training signal of residual prompts and learns useful information from them.

\textbf{Data Sampling Strategies.} The outputs of LLMs rely heavily on data sampling strategies to balance diversity and quality. Common strategies include greedy search, beam search, and various random sampling techniques such as top-k and top-p~\citep{zhao2023survey,minaee2024large}. In RLVR, the model generates on-policy responses and assigns them verifiable rewards during training. Basic random decoding is widely used in RLVR algorithms such as GRPO and DAPO~\citep{guo2025deepseek,yu2025dapo}. Beyond this, several works explore alternative sampling strategies. \citet{hou2025treerl} leverages tree search to find correct responses with higher probability. \citet{zheng2025first} forks responses at high-entropy tokens. \citet{shrivastava2025sample} dynamically allocates additional training resources to harder problems based on real-time difficulty estimates. \citet{xu2025not} selects a subset of responses to maximize reward variation. \citet{zheng2025act} predicts and skips uninformative prompts using reward training dynamics. \citet{zhang2025srpo} progressively exposes the model to increasingly challenging samples. Nevertheless, none of these methods are specifically designed to leverage information from residual prompts.
\section{Preliminaries}
\label{preliminaries}

\textbf{Notation}
We define an autoregressive language model parameterized by $\theta$ as a policy $\pi_\theta$. 
Let $q$ denote a query and $\mathcal{D}$ the query set. 
For a response $o$ to query $q$, its likelihood under $\pi_\theta$ is expressed as
\begin{equation}
\pi_\theta(o \mid q) = \prod_{t=1}^{|o|} \pi_{\theta}(o_{i,t} \mid q, o_{i,<t}),
\end{equation}
where $|o|$ is the number of tokens in $o$. 

\textbf{Group Relative Policy Optimization (GRPO)}~\citep{shao2024deepseekmath,guo2025deepseek} has shown strong effectiveness for fine-tuning LLMs. Unlike traditional approaches that rely on a critic network of comparable size to the policy, GRPO estimates the baseline directly from group-level rewards. For a specific question-answer pair $(q,a)$, the behavior policy $\pi_{\theta_\text{old}}$ samples a group of $G$ individual responses $\{ o_i\}_{i=1}^G$. Then, the advantage of the $i$-th response is calculated by normalizing the group-level rewards $\{ R_i \}_{i=1}^G$:
\begin{equation}
\hat{A}_{i,t} = \frac{r_i - \text{mean}(\{R_i\}_{i=1}^G)}{\text{std}(\{R_i\}_{i=1}^G)}.
\end{equation}

Building on the group-normalized advantages, GRPO optimizes the policy with a clipped objective that stabilizes updates and a KL regularization term that constrains divergence from the reference model. The objective is defined as:
\begin{equation}
\begin{aligned}
\mathcal{J}_\text{GRPO}(\theta)& = \mathbb{E}_{(q,a)\sim \mathcal{D}, \{o_i\}_{i=1}^G\sim \pi_{\theta_\text{old}}(\cdot\mid q)} \\&
\Bigg[ \frac{1}{G}\sum_{i=1}^{G} \frac{1}{|o_i|}\sum_{t=1}^{|o_i|} \Bigg( 
\min \Big( r_{i,t}(\theta) \hat{A}_{i,t},  
\ \text{clip} \Big( r_{i,t}(\theta), 1 - \varepsilon, 1 + \varepsilon \Big) \hat{A}_{i,t} \Big) \\ 
    &\quad - \beta D_{\text{KL}}(\pi_{\theta} || \pi_{\text{ref}}) 
\Bigg) \Bigg],
\label{eq:grpoloss}
\end{aligned}
\end{equation}
where $r_{i,t}(\theta)$ is the importance ratio between the old and new policy:
\begin{equation}
    r_{i,t}(\theta)=\frac{\pi_{\theta}(o_{i,t} \mid q, o_{i,<t})}{\pi_{\theta_{\text{old}}}(o_{i,t} \mid q,o_{i,<t})}.
\end{equation}

\textbf{Decoupled Clip and Dynamic sAmpling Policy Optimization (DAPO)}~\citep{yu2025dapo} introduces four key improvements: Clip-Higher promotes output diversity and mitigates entropy collapse; Dynamic Sampling is designed to enhance training efficiency and stability; Token-Level Policy Gradient Loss plays a critical role in handling long chain-of-thought reasoning; and Overlong Reward Shaping reduces reward noise while stabilizing optimization. Building on these components, DAPO optimizes the policy with the following objective:

\begin{equation}
\begin{aligned}
\mathcal{J}_{\text{DAPO}}(\theta) =\quad& \mathbb{E}_{(q,a)\sim \mathcal{D}, \{o_i\}_{i=1}^G\sim \pi_{\theta_\text{old}}(\cdot\mid q)}\\&
\Bigg[\frac{1}{\sum_{i=1}^{G}|o_i|}\sum_{i=1}^{G}\sum_{t=1}^{|o_i|} 
\min \Big( r_{i,t}(\theta) \hat{A}_{i,t},  
\ \text{clip} \Big( r_{i,t}(\theta), 1 - {\varepsilon_{\text{low}}}, 1 + {\varepsilon_{\text{high}}} \Big) \hat{A}_{i,t} \Big) \Bigg]
\\
\text{s.t.}\quad& 0< \Big|\{o_i\mid\texttt{is\_equivalent}(a,o_i)\}\Big|< G,
\label{eq:dapoloss}
\end{aligned}
\end{equation}
where $a$ is the ground-truth answer of query $q$.

\section{Methodology}
\label{methodology} 

\ERPO{} is proposed to leverage the training information contained in residual prompts to improve reinforcement learning for reasoning language models. Section~\ref{sec:reactivate} introduces the idea of reactivating the training signal from residual prompts that are otherwise discarded during RL training. Section~\ref{sec:ERPO} describes how \ERPO{} predicts whether a prompt is residual and adaptively modifies the sampling strategy. It further explains how \ERPO{} adjusts the sampling temperature to encourage different levels of exploration based on a history tracker.

\begin{figure}[t!]
  \centering
  \includegraphics[width=1\linewidth]{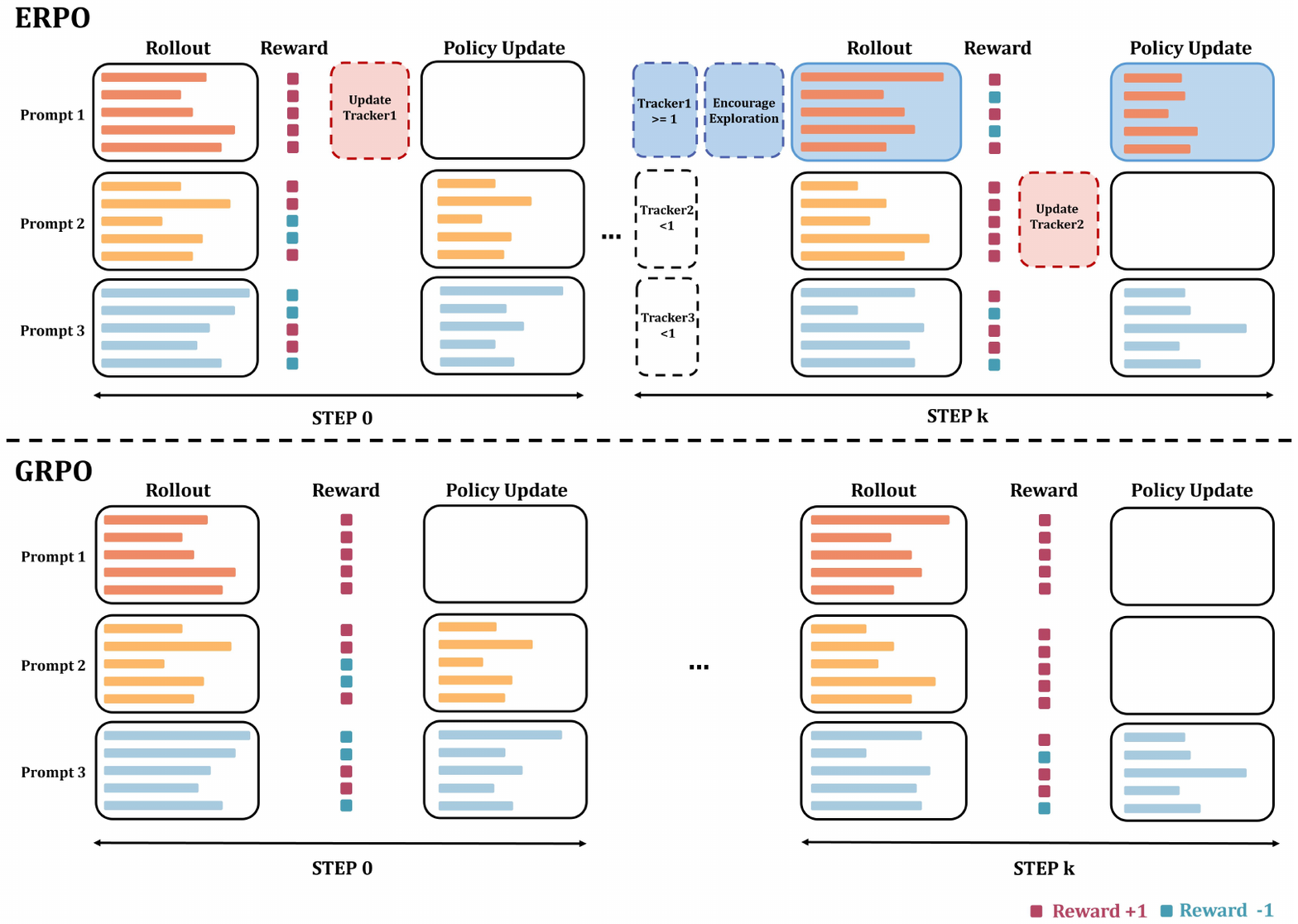} 
  \vspace{-15pt}
  \caption{ 
  Comparison between \ERPO{} and GRPO. During RL training, the policy gradually learns from the training data, resulting in more residual prompts with all-correct responses. In GRPO, residual prompts yield zero-variance rewards and thus provide no training signal for policy updates, reducing the effectiveness of training data. In contrast, \ERPO{} maintains a tracker for each prompt to record the number of times it produces all-correct responses, and adaptively encourages exploration on residual prompts to trigger incorrect responses and reactivate the training signal.
  }
  \label{fig:overview}
\end{figure}

\subsection{Reactivate Training Signal}
\label{sec:reactivate}
Current RLVR algorithms usually discard residual prompts that contain all-correct responses by assigning them zero advantage~\citep{guo2025deepseek} or by directly filtering them out from the training batch~\citep{yu2025dapo}. Consequently, available training prompts continually decreases with ongoing RL training and increasing model size, leading to a gradual reduction in both the size and diversity of the training dataset. As shown in Table~\ref{tab:gen_temp}, larger models with longer training generally produce more residual prompts that yield all-correct responses. Furthermore, as training progresses and the policy evolves, new reasoning traces and directions may be generated for the residual prompts, which can help the model learn more diverse reasoning patterns. In addition, training on residual prompts can reinforce the reasoning abilities the model has already acquired.

To leverage the training signal from residual prompts that are left behind by current RL algorithms, we propose a simple method to reactivate them. For residual prompts with all-correct responses, we replace the zero advantage with a small positive advantage by introducing a pseudo-negative reward into the advantage computation. The new Reactivated Advantage (RA) for prompts with all-correct responses is:
\begin{equation}
\hat{RA}_{i,t} =
\frac{r_{i} - \operatorname{mean}\!\left(\{R_{i}^{+}\}_{i=1}^{G} \cup \{R^{-}\}\right)}
{\operatorname{std}\!\left(\{R_{i}^{+}\}_{i=1}^{G} \cup \{R^{-}\}\right)}.
\end{equation}
where $R^{+}$ is the reward for correct responses and $R^{-}$ is the reward for incorrect responses. Using RA, residual prompts with all-correct responses still retain a small positive advantage, providing a valid training signal instead of being discarded by the RL algorithm.

\subsection{Explore Residual Prompts in Policy Optimization}
\label{sec:ERPO}

Although using the reactivated advantage can force the model to learn information from residual prompts, residual prompts will dominate the trainig along the training process and model scales up, leaving less negative feedback and impede the training effectiveness.~\citep{chen2025bridging,xiong2025minimalist}. Furthermore, the model may suffer from the imbalance between exploration and exploitation, overfitting to a narrow exploration space.

To address this limitation, we propose to adaptively encourage exploration on residual prompts by controling their sampling temperature. As shown in Table~\ref{tab:gen_temp}, a higher sampling temperature can trigger incorrect responses, thereby reactivating the training signals of residual prompts. Note that training data typically exhibit strong temporal correlations across epochs~\citep{zheng2022coverage}, meaning that a prompt producing all-correct responses in the current epoch is likely to do so again in the following epoch~\citep{zheng2025act}. Thus, we can maintain a history tracker $H_i$ to track how many times the policy generates all-correct responses for a prompt $q_i$:
\begin{equation}
H_i^{(0)} = 0,\qquad
H_i^{(t)} = H_i^{(t-1)} + \1_\mathrm{\text{$q_i$ has all-correct responses at step $t$}}
\end{equation}
Then $H_i$ is used to determine whether we should assign a larger sampling temperature to prompt $q_i$. If $H_i$ is greater than $0$, it means that prompt $q_i$ is already easy for the policy to generate all-correct responses, and it is very likely to provide no training signals the next time the policy samples it. Therefore, we assign a larger sampling temperature to prompts with $H_i > 0$ to encourage more exploration of their reasoning traces and to reactivate the training signal by triggering incorrect responses.

Since the robustness of prompts varies, some residual prompts require only a marginal increase in sampling temperature to induce incorrect responses, whereas others necessitate substantially larger adjustments. At the same time, it is essential to preserve the benefits of on-policy learning by constraining distributional shifts within a reasonable range to ensure stable and effective training. Assigning excessively large sampling temperatures is particularly detrimental for prompts with lower robustness. This trade-off highlights the difficulty of selecting a single, unified sampling temperature that can consistently induce incorrect responses, enhance exploration, and maintain a manageable distribution shift across all residual prompts. Therefore, \ERPO{} introduces a prompt-adaptive adjustment of the sampling temperature:
\begin{equation}
T_i^{(t)} = \min(T_0 + T_s \cdot H_i^{(t)}, T_{max})
\end{equation}
where $T_0$, $T_{max}$, and $T_s$ are hyperparameters representing the initial temperature, maximum temperature, and temperature step size, respectively. In this way, \ERPO{} gradually increases the sampling temperature of residual prompts until the policy generates incorrect responses. This enables \ERPO{} to strike a balance between reactivating training signals, encouraging exploration, and maintaining a reasonable distribution shift. In general, our \ERPO{} framework can be summarized in Algorithm~\ref{alg:edlb}:

\begin{algorithm}[t]
\caption{\ERPO{} framework}
\label{alg:edlb}
\textbf{Input:} Policy $\pi_\theta$, reward model $R$,\\
\hspace*{3.2em} Prompt set $\mathcal{D}=\{q_i\}_{i=1}^N$, history tracker $\{H_i\}_{i=1}^N$ (init.\ $H_i^{(0)}{=}0$),\\
\hspace*{3.2em} Rollouts per prompt $n$, temperatures $(T_0, T_{\max})$, step size $T_s$, steps $K$\\[2pt]
\textbf{Output:} Updated policy $\pi_{\theta_{\mathrm{updated}}}$
\vspace{3pt}
\begin{algorithmic}[1]
\For{$t = 1,2,\dots,K$}
  \State Sample a mini-batch $\mathbf{q} \subseteq \mathcal{D}$ 
  \For{each $q_i \in \mathbf{q}$}
    \State $T_i^{(t)} \leftarrow \min\!\big(T_0 + T_s \cdot H_i^{(t-1)},\, T_{\max}\big)$
    \State Sample $n$ rollouts $\mathbf{o_i}=(o_1,\dots,o_n)$ for $q_i$ using $\pi_\theta$ at temperature $T_i^{(t)}$
    \State Compute rewards $\mathbf{r_i}=(r_1,\dots,r_n)$ with $R$; \ \ $\mathsf{acc}\leftarrow \1_\mathrm{\text{all $o_j$ correct}}$
    \State  $H_i^{(t)} \leftarrow H_i^{(t-1)} + \1_\mathrm{\mathsf{acc}=1}$
  \EndFor
  \State Update $\pi_\theta$ using an RL algorithm with data $\mathcal{B}=\{(\mathbf{q},\mathbf{o},\mathbf{r})\}$
\EndFor
\State \textbf{return} $\pi_{\theta_{\mathrm{updated}}} \leftarrow \pi_\theta$
\end{algorithmic}
\end{algorithm}

\section{Experiments}
\label{experiments}

In this section, we first outline the implementation details, including training details and evaluation. We then present the main results, comparing \ERPO{} against baseline approaches across several math reasoning benchmarks. Finally, we provide additional experimental results to support further analysis.

\subsection{Implementation Details}
\label{sec:implementation}

\textbf{Training details:} Following recent studies~\citep{zheng2025first,cheng2025reasoning,shao2025spurious} that apply RLVR to train LMMs for math reasoning tasks, we adopt Qwen2.5-3B and Qwen2.5-7B~\citep{qwen2025qwen25technicalreport} as our backbone models. Consistent with prior work~\citep{yu2025dapo,cheng2025reasoning,cui2025entropy}, we use the DAPO-Math-17K dataset~\citep{yu2025dapo} for training. To achieve strong performance, we adopt the DAPO algorithm~\citep{yu2025dapo}. Prior works~\citep{yu2025dapo,cheng2025reasoning} has demonstrated its superior effectiveness and stability over vanilla GRPO, and we employ it both as the baseline and as the optimization method for \ERPO{}. The learning rate is set to $1 \times 10^{-6}$ with a linear warm-up over 10 rollout steps. For rollout, we use a prompt batch size of 512, sampling 16 responses per prompt. During training, the mini-batch size is set to 512, resulting in 16 gradient updates per rollout step. The initial rollout temperature $T_0$ is set to $1.0$. The temperature increment step $T_s$ is set to $0.02$ for Qwen2.5-3B and $0.05$ for Qwen2.5-7B, while the maximum rollout temperature $T_{\max}$ is set to $1.2$ for Qwen2.5-3B and $1.4$ for Qwen2.5-7B, respectively. Rewards are assigned as $1$ for correct responses and $-1$ otherwise. All experiments are conducted using the verl framework~\citep{sheng2024hybridflow}. More details can be found in the Appendix.

\noindent\textbf{Evaluation:} We evaluate our models on AIME 2025/2024, AMC 2023, and MATH500~\citep{hendrycks2021measuring}, using a rollout temperature of $1.0$ and top-$p$ sampling with $p = 0.7$. For AIME and AMC, we sample $32$ independent responses for each prompt and report the average accuracy as $mean@32$. In addition, we provide the majority-vote~\citep{zhao2023survey}  accuracy $maj@32$ as a complementary metric. For the larger and less challenging MATH500 benchmark, we sample $4$ responses per prompt and report the average accuracy $mean@4$ and majority-vote accuracy $maj@4$. All evaluations are conducted using the verl framework~\citep{sheng2024hybridflow} and follow the same evaluation protocol as DAPO~\citep{yu2025dapo}. More details can be found in the Appendix.

% ----------------------------------------------------------------------

\subsection{Benchmark Comparisons}
\label{main_exp}

In this section, we compare the performance of DAPO, Reactivated Advantage (RA), and \ERPO{} on the AIME25, AIME24, AMC23, and MATH500 benchmarks. The detailed results are shown in Table~\ref{tab:main_results}. On Qwen-3B, both RA and \ERPO{} achieve higher mean and majority-vote accuracy than the baseline DAPO. The improvement of RA demonstrates that residual prompts still contain valuable training information and should not be totally excluded from RL training. On AIME2025, RA achieves a remarkable performance gain compared to the baseline: around a $70\%$ improvement on $mean@32$ and a $28\%$ improvement on $maj@32$. Since AIME2025 is shown to suffer less from data contamination during model pretraining than the other math benchmarks~\citep{wu2025reasoning}, these results confirm that learning on residual prompts is particularly helpful for tasks that are novel and challenging for the model.

On Qwen2.5-7B, \ERPO{} achieves the best overall performance in both mean and majority-vote accuracy compared to DAPO and RA, indicating the scalability of our algorithm. On AIME2025, \ERPO{} achieves the largest improvement over the baseline, with an increase of approximately $12\%$ and $16\%$ on $mean@32$ and $maj@32$. However, unlike the results on the 3B model, RA performs worse than \ERPO{}. A possible reason is that reactivating all residual prompts may lead to overfitting when the proportion of residual prompts is high during training. As shown in Table~\ref{tab:gen_temp}, Qwen2.5-7B has more than $20\%$ residual prompts, making this issue more pronounced as the model scales up. In contrast, \ERPO{} avoids this problem by setting $T_{\max}$, which prevents unbounded increases in sampling temperature. Once a residual prompt is fully learned and robust to higher temperatures, it no longer provides a training signal.

In summary, RA verifies that residual prompts contain information that can still benefit model training and should not be totally discarded. \ERPO{} further provides an effective sampling strategy that leverages training signals from residual prompts and scales effectively to larger models.

\begin{table}[t!]
\centering
\caption{Performance comparison of the Qwen2.5-3B and Qwen2.5-7B models trained with DAPO, DAPO+RA, and DAPO+\ERPO{}. Evaluations use mean@32 and maj@32 for AIME25, AIME24 and AMC23; MATH500 is reported with mean@4 and maj@4. The Avg. columns average the mean and maj across datasets.}
\small
\setlength{\tabcolsep}{3pt}
\resizebox{\textwidth}{!}{%
\begin{tabular}{lcccccccccc}
\toprule
\multirow{2}{*}{\textbf{Method}}
& \multicolumn{2}{c}{\textbf{AIME25}}
& \multicolumn{2}{c}{\textbf{AIME24}}
& \multicolumn{2}{c}{\textbf{AMC23}}
& \multicolumn{2}{c}{\textbf{MATH500}}
& \multicolumn{2}{c}{\textbf{Avg.}} \\
\cmidrule(lr){2-3}\cmidrule(lr){4-5}\cmidrule(lr){6-7}\cmidrule(lr){8-9}\cmidrule(lr){10-11}
& mean@32 & maj@32
& mean@32 & maj@32
& mean@32 & maj@32
& mean@4  & maj@4
& mean    & maj \\
\midrule
\multicolumn{11}{c}{Qwen2.5-3B} \\
\midrule
\textit{DAPO} & 4.5 & 8.9 & 9.5 & 15.2 & 58.4 & 68.0 & 50.4 & 50.8 & 30.7 & 35.7 \\
+RA           &  \textbf{7.7}   &  \textbf{11.4}   &  9.6   &  14.1  &  59.1  &  67.1  & \textbf{52.0} & \textbf{52.5} & \textbf{32.1} & 36.3 \\
+\ERPO{}      & 5.5 & 8.8 & \textbf{10.3} & \textbf{16.0} & \textbf{60.8} & \textbf{70.0} & 51.1 & 51.6 & 31.9 & \textbf{36.6} \\
\midrule
\multicolumn{11}{c}{Qwen2.5-7B} \\
\midrule
\textit{DAPO} & 12.6 & 16.9 & 17.5 & 20.1 & \textbf{76.7} & 81.4 & 60.3 & 60.9 & 41.8 & 44.8 \\
+RA           & 13.5 & 16.4 & 16.1 & 18.1 & 75.8 & 80.2 & 61.2 & 61.6 & 41.7 & 44.1 \\
+\ERPO{}      & \textbf{14.2} & \textbf{19.4} & \textbf{19.0} & \textbf{21.2} & 76.4 & \textbf{81.5} & \textbf{61.7} & \textbf{62.0} & \textbf{42.8} & \textbf{46.0} \\
\bottomrule
\end{tabular}}
\label{tab:main_results}
\end{table}

% ----------------------------------------------------------------------

\subsection{Exploration on Residual Prompts}

To investigate the effect of sampling temperature on residual prompts, we conduct experiments to measure the proportion of residual prompts under different temperature settings. Specifically, we select a $2$k subset from our training dataset DAPO-Math-17K, sample each prompt $16$ times following the same training configuration, and calculate the proportion of residual prompts within this subset. We evaluate Qwen2.5-3B, Qwen2.5-7B, and Qwen2.5-32B trained with DAPO. For Qwen2.5-32B+DAPO, we use the publicly released checkpoints from DAPO~\citep{yu2025dapo}. The detailed results are presented in Table~\ref{tab:gen_temp}. Our findings highlight three key observations: (1) the proportion of residual prompts increases after training; (2) larger models tend to produce more residual prompts, revealing the challenge of scaling RLVR with model size; and (3) higher sampling temperatures encourage greater exploration and can elicit more incorrect responses from residual prompts.

On the other hand, we also examine the effect of sampling temperature on prompts with all-incorrect responses. The experimental settings are kept the same, and the results are presented in Table~\ref{tab:incorrect}. The findings indicate that RL training and model scaling reduce the proportion of prompts with all-incorrect responses. Moreover, sampling temperature has a much smaller impact on this proportion than on residual prompts. Therefore, \ERPO{} is applied only to residual prompts that are more likely to yield all-correct responses.

\begin{table}[t!]
\centering
\caption{Proportion of prompts with \textbf{all-incorrect} responses under different sampling temperatures and model scales.}
\setlength{\tabcolsep}{4pt}
\renewcommand{\arraystretch}{1.2}
\begin{tabular}{lccc}
    \toprule
      & {\textbf{$T=1.0$}} & {\textbf{$T=1.1$}} & {\textbf{$T=1.2$}} \\
    \midrule[0.6pt]
    { Qwen2.5-3B }   & $73.0\%$  & -- & --  \\
    { Qwen2.5-3B + DAPO}   & $39.6\%$  & $40.6\%$ & $44.0\%$    \\
    { Qwen2.5-7B }   & $68.9\%$ & -- & --  \\
    { Qwen2.5-7B + DAPO}   & $25.1\%$ & $26.9\%$ & $33.2\%$    \\
    { Qwen2.5-32B }  & $48.9\%$ & -- & --  \\
    { Qwen2.5-32B + DAPO}  & $7.0\%$ & $7.9\%$ & $ 15.2\%$    \\
    \bottomrule
\end{tabular}
\vspace{-6pt}
\label{tab:incorrect}
\end{table}

% --------------------------------------

\subsection{Sensitive Analysis}

We conduct a sensitivity analysis on the temperature increment step ($T_s$) and the maximum rollout temperature ($T_{\max}$) to evaluate their impact on the Qwen2.5-3B model. Specifically, we experiment with three parameter settings: $T_s=0.02, T_{\max}=1.2$; $T_s=0.05, T_{\max}=1.2$; and $T_s=0.05, T_{\max}=1.4$. The performance under these settings is reported in Table~\ref{tab:sensitive}. The results show that $T_s=0.02, T_{\max}=1.2$ yields the best performance, while increasing either $T_s$ or $T_{\max}$ leads to performance degradation. Notably, the setting $T_s=0.05, T_{\max}=1.2$ still outperforms the vanilla DAPO algorithm, whereas $T_s=0.05, T_{\max}=1.4$ achieves performance comparable to the DAPO baseline. Since smaller models are generally less robust than larger ones, these results on the 3B model further validate the effectiveness of \ERPO{} and demonstrate that it is not highly sensitive to the choice of hyperparameters.

\begin{table}[t!]
\centering
\caption{Sensitivity analysis on temperature increment step $T_s$ and maximum rollout temperature $T_{\max}$ using Qwen2.5-3B. Evaluations use mean@32/maj@32 for AIME25, AIME24, AMC23; MATH uses mean@4 and maj@4. The Avg. columns average mean and maj across datasets (using mean@4/maj@4 for MATH).}
\small
\setlength{\tabcolsep}{3pt}
\resizebox{\textwidth}{!}{%
\begin{tabular}{lcccccccccc}
\toprule
\multirow{2}{*}{\textbf{Setting}}
& \multicolumn{2}{c}{\textbf{AIME25}}
& \multicolumn{2}{c}{\textbf{AIME24}}
& \multicolumn{2}{c}{\textbf{AMC23}}
& \multicolumn{2}{c}{\textbf{MATH}}
& \multicolumn{2}{c}{\textbf{Avg.}} \\
\cmidrule(lr){2-3}\cmidrule(lr){4-5}\cmidrule(lr){6-7}\cmidrule(lr){8-9}\cmidrule(lr){10-11}
& mean@32 & maj@32
& mean@32 & maj@32
& mean@32 & maj@32
& mean@4  & maj@4
& mean    & maj \\
\midrule
{\scriptsize $T_s=0.02, T_{\max}=1.2$} & 5.5 & 8.8 & 10.3 & 16.0 & 60.8 & 70.0 & 51.1 & 51.6 & 31.9 & 36.6  \\
{\scriptsize $T_s=0.05, T_{\max}=1.2$} & 5.6 & 9.4 & 8.8 & 14.2 & 59.2 & 69.5 & 51.2 & 51.9 & 31.2 & 36.3 \\
{\scriptsize $T_s=0.05, T_{\max}=1.4$} & 3.6 & 5.5 & 9.9 & 14.6 & 59.6 & 67.7 & 52.2 & 52.7 & 31.3 & 35.1 \\
\bottomrule
\end{tabular}}
\label{tab:sensitive}
\end{table}

% --------------------------------------

\subsection{Further Analysis}

\begin{figure}[t!]
  \centering
  \includegraphics[width=1\linewidth]{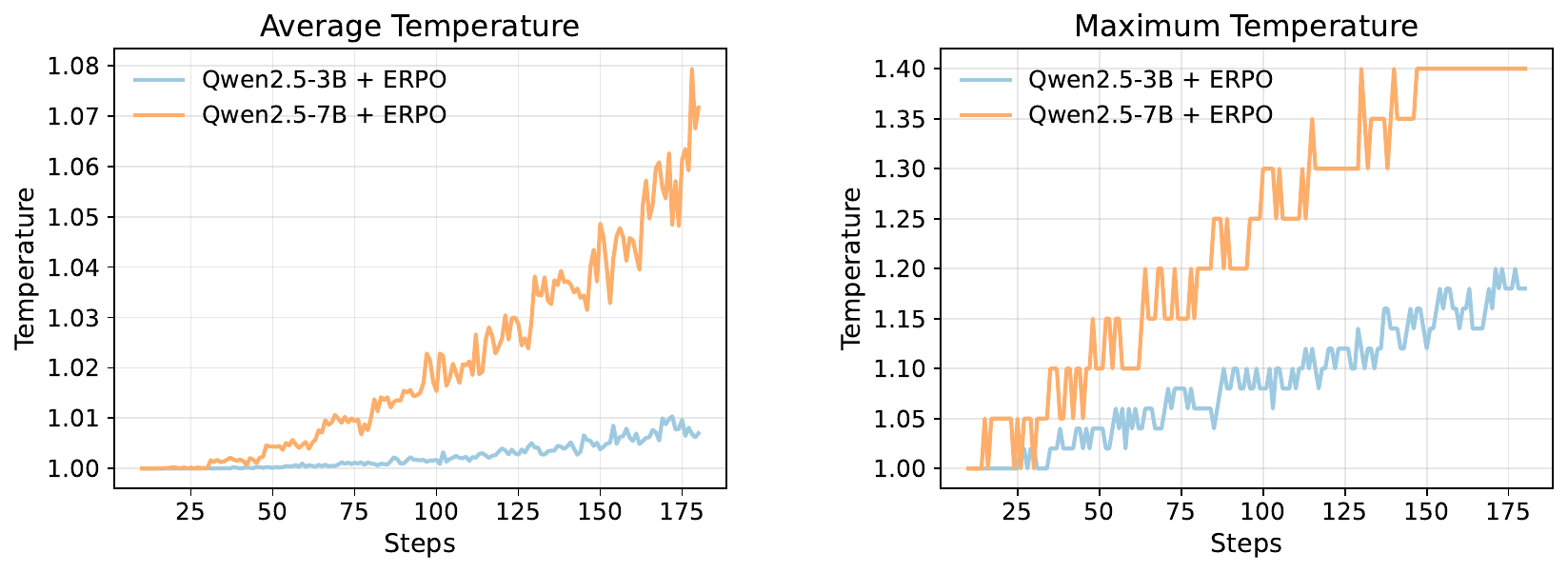} 
  \vspace{-15pt}
  \caption{ The average and maximum sampling temperatures during the \ERPO{} training process. The steps shown here are the prompt generation steps.
  }
  \label{fig:temperature}
\end{figure}

\textbf{Sampling Temperature} The average and maximum sampling temperatures during the \ERPO{} training process are shown in Figure\ref{fig:temperature}. The maximum temperature increases linearly, whereas the average temperature increases exponentially, indicating that more prompts become residual prompts whose sampling temperatures are raised by \ERPO{}. Setting an upper bound on the temperature, $T_{\max}$, is necessary to prevent uncontrolled growth of the sampling temperature.

\begin{figure}[t!]
  \centering
  \includegraphics[width=1\linewidth]{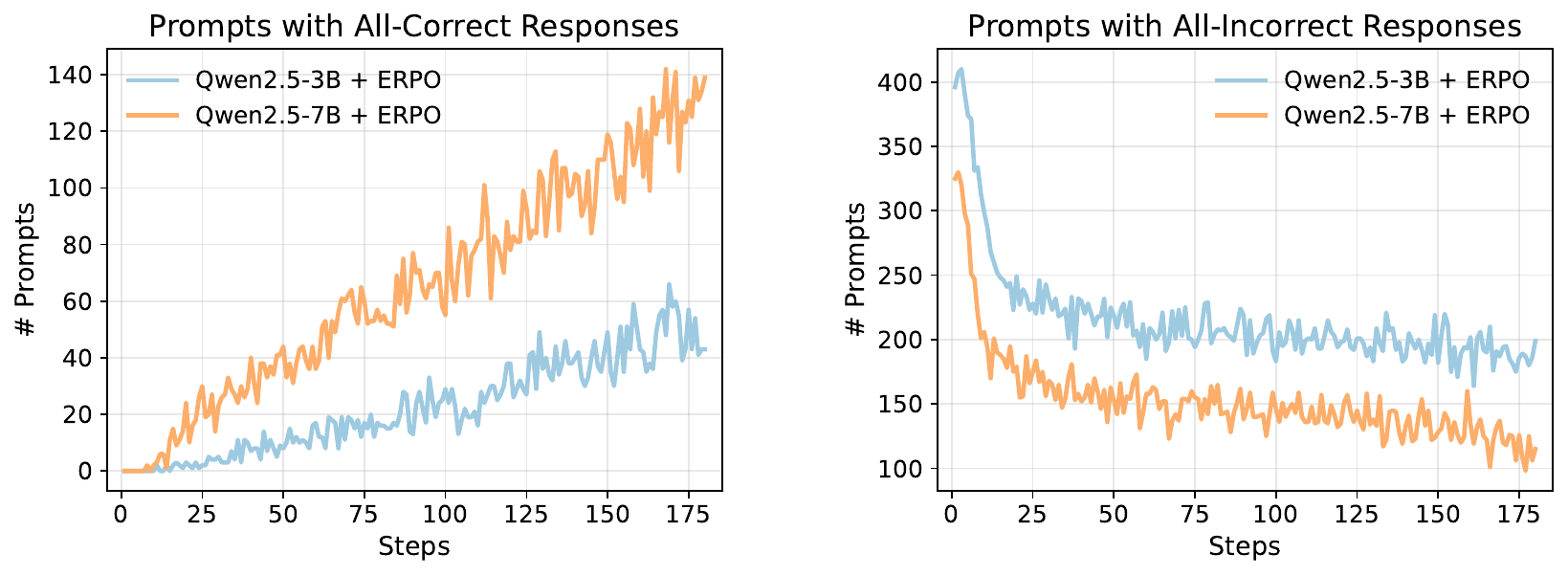} 
  \vspace{-15pt}
  \caption{ The number of residual prompts with all-correct responses and prompts with all-incorrect responses during the \ERPO{} training process. The steps shown here are the prompt generation steps.
  }
  \label{fig:prompts}
\end{figure}

\textbf{Residual Prompts} Figure~\ref{fig:prompts} shows the number of residual prompts with all-correct responses and the number of prompts with all-incorrect responses in a training batch of size $512$. During training, the number of prompts with all-incorrect responses continues to decrease, while the number of residual prompts steadily increases. Moreover, the growth rate of residual prompts is higher than the decay rate of all-incorrect prompts, underscoring the importance of leveraging residual prompts. In addition, the history tracker for Qwen2.5-3B and Qwen2.5-7B indicates that $15.3\%$ and $40.2\%$ of the prompts in the training dataset have a record $H_i > 0$, further demonstrating the critical role of \ERPO{} in the training process.

\section{Conclusion}
In this work, we address a key limitation of GRPO-based reinforcement learning for LLMs: the accumulation of residual prompts that diminish training diversity and leave valuable signals underutilized. To tackle this, we introduce the \ERPO{} framework, which adaptively adjusts sampling temperature based on prompt history to reactivate training signals and encourage broader exploration. Our experiments across multiple math reasoning benchmarks demonstrate that \ERPO{} not only mitigates prompt collapse but also improves both average and majority-vote performance, with especially strong gains on tasks less affected by data contamination. These results highlight the potential of exploiting residual prompts as a promising direction for advancing reinforcement learning with verifiable rewards.

\section*{Ethics statement}
This work uses only publicly available mathematical datasets without personal or sensitive information. The study does not involve human subjects or animals. Our method focuses on improving reasoning in math tasks, with minimal risk of societal harm, and is intended solely for research purposes.

% \section*{Reproducibility statement}
% To facilitate reproducibility, we provide detailed implementation settings in Section~\ref{sec:implementation} and section~\ref{appendix:implementation} in the Appendix. In addition, we release the source code in the supplementary materials, enabling readers to replicate all experiments and results reported in this paper.



\clearpage
\bibliography{iclr2026_conference}
\bibliographystyle{iclr2026_conference}

\clearpage
\appendix
\begin{center}
    {\LARGE \bf Appendix}
\end{center}

\section{Experiemntal Details}
\label{appendix:implementation}
\subsection{Training Details}
We provide detailed settings of various parameters in the DAPO algorithm, which serves as both the baseline and the optimization method for Reactivated Advantage (RA) and \ERPO{}. The KL coefficient is fixed at $0$ across all experiments. The clip ratio is set to $\epsilon_{low}=0.2$ and $\epsilon_{high}=0.28$. The maximum response length is set to $10{,}240$ for experiments on the Qwen2.5-3B model and the RA algorithm, and to $20{,}480$ for experiments with DAPO and \ERPO{} on the Qwen2.5-7B model. The overlong buffer is set to $4{,}096$, with an overlong penalty factor of $1$. $180$ prompt generation steps/ $2880$ policy update steps is used for all emperiments.

\subsection{Evaluation Details}
We follow the same evaluation protocol as DAPO~\citep{yu2025dapo}, using the verl framework~\citep{} to assess all benchmarks. Specifically, each question from the benchmark is prepended with the prompt \texttt{Solve the following math problem step by step. The last line of your response should be of the form Answer: \$Answer (without quotes), where \$Answer is the solution to the problem.\textbackslash n\textbackslash n} and appended with the prompt \texttt{\textbackslash n\textbackslash nRemember to put your answer on its own line after "Answer:"}. This structure is identical to that used in the training data. We then follow the same workflow as DAPO to extract the final answer from the model responses.

\section{Additional Experimental Results} 
\textbf{Comparison with additional baselines.} We futher compare our model performance with another baseline Entropy~\citep{cheng2025reasoning}, which use the same backbone model Qwen2.5-7B, training dataset DAPO-Math-17K, optimization method DAPO and very similiar hyperparameters. The results are shown in Table~\ref{tab:appendix_results}. Results show that \ERPO{} outperforms Entropy on every benchmark by a large margin, demonstrating the effectiveness of \ERPO{}. 

\begin{table}[!htbp]
\centering
\caption{Performance comparison of the Qwen2.5-7B model trained with the Entropy baseline and \ERPO{}. For the Entropy baseline, we report the values provided in their paper. For \ERPO{}, we report the $mean@32$ scores on AIME25, AIME24, and AMC23, and the $mean@4$ score on MATH500.}
\small
\vspace{2pt}
\setlength{\tabcolsep}{3pt}
\begin{tabular}{lccccc}
\toprule
\textbf{Method} & \textbf{AIME25} & \textbf{AIME24} & \textbf{AMC23} & \textbf{MATH500} & \textbf{Avg.} \\
\midrule
\multicolumn{6}{c}{Qwen2.5-7B} \\
\midrule
Entropy   & 11.8 & 12.6 & 57.8 & 58.5 & 35.2 \\
ERPO        & 14.2 & 19.0 & 76.4 & 61.7 & 42.8 \\
\bottomrule
\end{tabular}
\label{tab:appendix_results}
\end{table}

\begin{figure}[h]
  \centering
  \includegraphics[width=1\linewidth]{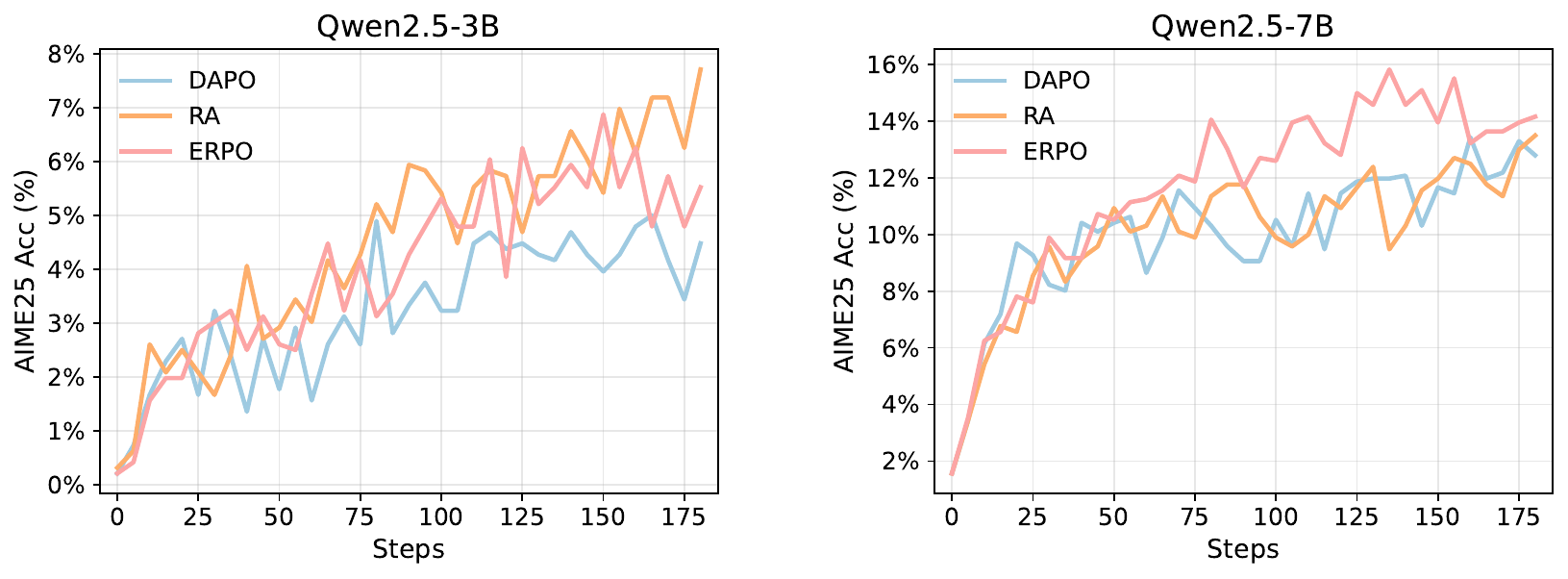} 
  \vspace{-15pt}
  \caption{Performance of $mean@32$ on AIME2025. 
  }
  \label{fig:aime}
\end{figure}

\textbf{Model Performance.} We further present the model performance on AIME25 throughout training in Figure~\ref{fig:aime}. On both Qwen2.5-3B and Qwen2.5-7B, \ERPO{} consistently outperforms the DAPO baseline for most of the training process, demonstrating its effectiveness on novel and challenging math tasks that are less affected by data contamination. RA achieves the best performance on Qwen2.5-3B but only the second-best on Qwen2.5-7B, suggesting that training on residual prompts can provide notable benefits, but its advantages diminish as model size scales up.

\end{document}